\documentclass{article}
\usepackage[hidelinks]{hyperref}

\usepackage{spconf,amsmath,epsfig}
\usepackage[noadjust]{cite}
\usepackage{amssymb}
\usepackage{amsfonts}
\usepackage{mathtools}
\usepackage{bbm}
\usepackage{pifont}

\usepackage{upgreek}
\usepackage{float}
\usepackage{subcaption}
\usepackage{multicol}
\usepackage{multirow}
\usepackage{verbatim}
\usepackage{soul}
\usepackage{flushend}

\usepackage{amsmath,amssymb,amsfonts}
\usepackage[linesnumbered,lined,boxed,commentsnumbered,ruled]{algorithm2e}
\usepackage[usenames,dvipsnames]{xcolor} 
\usepackage{orcidlink}

\usepackage{adjustbox}
\usepackage{makecell}
\usepackage{comment}
\usepackage{pbalance}

\DeclareMathAlphabet\mathbfcal{OMS}{cmsy}{b}{n}
\title{Communication-Efficient Federated Learning based on Explanation-Guided Pruning for Remote Sensing Image Classification}
\name{Jonas Klotz$^*$ \orcidlink{0009-0003-3297-2365}, Bar{ı}\c{s} B\"{u}y\"{u}kta\c{s}$^*$ \orcidlink{0000-0002-8240-8784}, Beg\"{u}m Demir \orcidlink{0000-0003-2175-7072}} 
\address{
BIFOLD - Berlin Institute for the Foundations of Learning and Data, Germany\\
Faculty of Electrical Engineering and Computer Science, Technische Universit\"at Berlin, Germany
}

\begin{document}
%
\maketitle
\def\thefootnote{*}\footnotetext{These authors contributed equally to this work}

\begin{abstract}

Federated learning (FL) is a decentralized machine learning paradigm in which multiple clients collaboratively train a global model by exchanging only model updates with the central server without sharing the local data of the clients. Due to the large volume of model updates required to be transmitted between clients and the central server, most FL systems are associated with high transfer costs (i.e., communication overhead). This issue is more critical for operational applications in remote sensing (RS), especially when large-scale RS data is processed and analyzed through FL systems with restricted communication bandwidth. To address this issue, we introduce an explanation-guided pruning strategy for communication-efficient FL in the context of RS image classification. Our pruning strategy is defined based on the layer-wise relevance propagation (LRP) driven explanations to: 1) efficiently and effectively identify the most relevant and informative model parameters (to be exchanged between clients and the central server); and 2) eliminate the non-informative ones to minimize the volume of model updates. The experimental results on the BigEarthNet-S2 dataset demonstrate that our strategy effectively reduces the number of shared model updates, while increasing the generalization ability of the global model. The code of this work is publicly available at https://git.tu-berlin.de/rsim/FL-LRP.
\end{abstract}
\section{Introduction}
Federated learning (FL) has emerged as a promising paradigm for training machine learning models collaboratively, while ensuring data privacy. In detail, it allows training a deep learning (DL) model without having direct access to training data distributed across decentralized archives (i.e., clients), particularly in the case that data is unshared due to commercial concerns, privacy constraints, or legal regulations \cite{buyuktas2024federated}. To this end, each client independently trains a local model using the data on the clients, and then computes and sends only the model updates (e.g., weights or gradients) to a central server (i.e., a global model). The central server aggregates these updates to adjust the global model, which is then sent back to the clients for further training. This approach ensures data privacy, while allowing model training across clients. The development of FL methods has gained significant attention in remote sensing (RS) \cite{Buyuktas:2023,buyuktacs2024transformer,moreno2024federated,zhu2023privacy}. We refer the reader to \cite{buyuktas2024federated} for the details on FL in general and the analysis of the state-of-the-art FL methods in RS in particular.

During FL, the iterative exchange of model parameters between clients and the central server often involves transmitting large amounts of model updates. In particular, for operational RS applications, the communication overhead (i.e., transmission cost) becomes a critical constraint in the case of the presence of: 1) a high number of training samples at each client; 2) a high number of participating clients; 3) limited bandwidth; 4) high latency; and 5) energy-constrained clients \cite{wang2022communication}. All these issues make the deployments of FL systems impractical \cite{luping2019cmfl}, limiting their applicability in RS. The existing FL methods in RS do not address these issues, whereas communication-efficient FL has been widely studied in the machine learning (ML) and computer vision (CV) communities. Existing methods can be grouped into three categories: 1) model compression \cite{zhu2023model,wu2023fedcomp}; 2) knowledge distillation \cite{chen2023resource,yao2023f}; or 3) pruning \cite{jiang2022model, kumar2024fednisp}. The model compression-based methods aim to minimize communication overhead by compressing model parameters to be transmitted to the central server. However, they introduce additional computational overhead on both the clients and the central server due to the need for compressing and decompressing model updates. 
The knowledge distillation-based methods transfer knowledge from a global model to smaller models on clients. These methods impose significant computational demands as they require teacher-student training, involving the calculation of complex loss functions and the transfer of intermediate representations. The pruning-based methods aim to reduce communication overhead by removing less significant parameters. Unlike model compression and knowledge distillation-based methods, pruning-based methods introduce lower computational overhead. 

We focus our attention on pruning-based methods and investigate the effectiveness of explanation-based pruning in the context of FL in RS for the first time. In particular, we introduce a strategy that adapts the layer-wise relevance propagation (LRP)-based pruning technique proposed in \cite{yeom2021pruning} for FL in RS. Our strategy removes the least significant parameters at the central server, preserving the generalization capability of models, while minimizing communication overhead.

\begin{figure*}[h]
\centering
\centerline{\includegraphics[width=1.85\columnwidth,keepaspectratio]{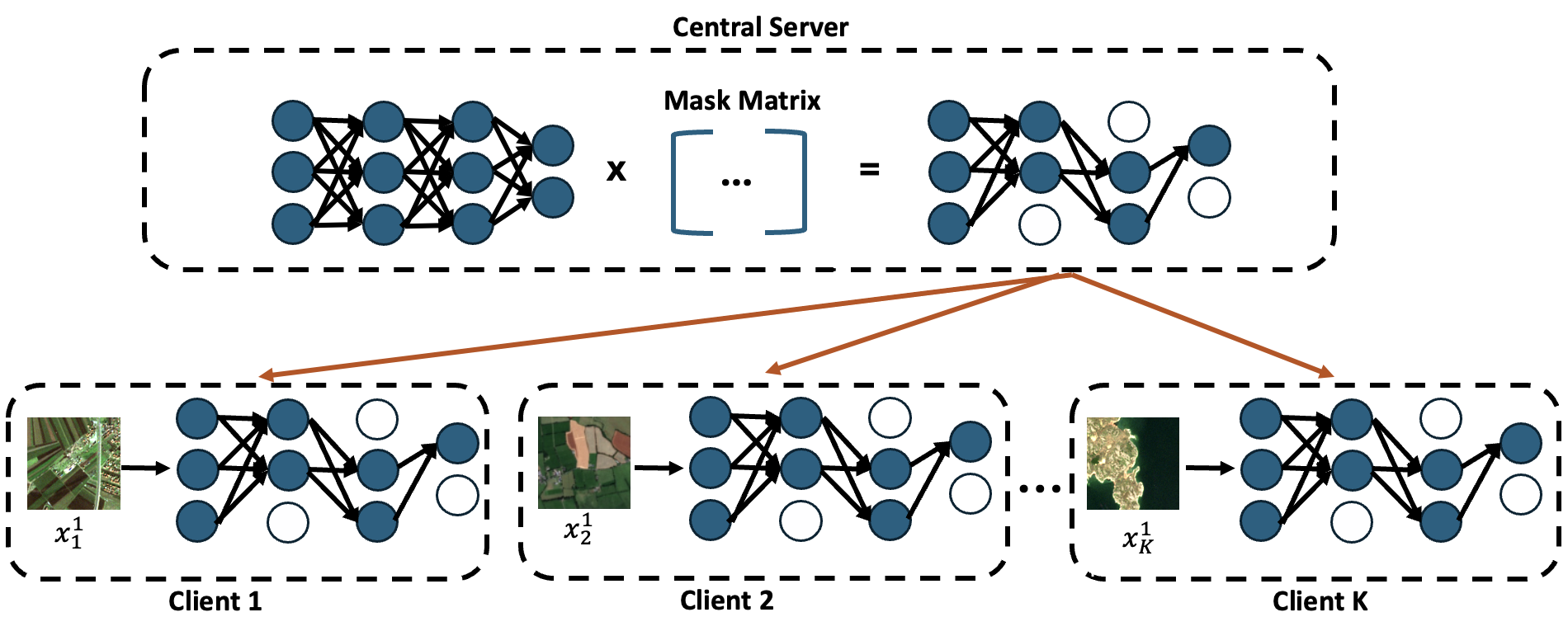}}
\caption{An illustration of the pruning process and distribution of the pruned model to clients in the proposed strategy.}
\label{figure:pruning}
\end{figure*}



    
 
        
        


\section{Proposed Explanation-Guided Pruning Strategy for FL}

Let us assume that a set of $K$ clients $\{C_1, C_2,...,C_K\}$ is available, where $C_i$ is the $i$-th client and $K$ is the total number of clients. $C_i$ has $M_i$ pairs $D_i = \{(\boldsymbol{x}_{i}^z, \boldsymbol{y}_{i}^z)\}_{z=1}^{M_i}$, where $\boldsymbol{x}_{i}^z$ is the $z$-th training image and $\boldsymbol{y}_{i}^z$ is the associated class annotation. For scene-level annotations, each training image is labeled with either a single label or multiple labels that correspond to the overall content of the scene. We assume that the data on clients is not shared, $M_i$ is very large, and the communication bandwidth available to the clients is limited. Thus, the communication overhead resulting from transmitting large volumes of model updates to the central server becomes a critical challenge. To address this issue, we introduce an explanation-guided pruning strategy applied at the central server in the context of FL. 

As the definition of the proposed strategy strictly depends on FL, the detailed explanation of the proposed strategy starts from the brief formulation of FL. Let us assume that the local DL model $\phi_i$ is trained using $D_i$ for $E$ epochs. Each client $C_i$ aims to find the optimal local model parameters $w_i$ by minimizing the local objective $\mathcal{O}_i$ as follows:
\begin{equation}
\begin{aligned}
\mathcal{O}_i (\mathcal{B};w_i) &=\!\!\!\!\!\!\! \sum_{(\boldsymbol{x}_i^z,\boldsymbol{y}_i^z)\in \mathcal{B}}\!\!\!\!\mathcal{L}(\phi_i(\boldsymbol{x}_{i}^z;w_i), \boldsymbol{y}_{i}^{z}),\\ 
w_i^* &= \arg \min_{w_i} \mathcal{O}_i(D_i;w_i),
\end{aligned}
\end{equation}
where $\mathcal{L}$ represents the task-specific loss function, such as categorical cross-entropy (CE) for scene-level single-label classification, or binary CE for scene-level multi-label classification (MLC). FL methods collaboratively learn the parameters $w^*$ of the global model $\phi^*$ over the entire training set $D_\text{train} = \bigcup_{i \in \{ 1, 2,..., K \} } D^i$, by minimizing the following objective:
\begin{equation}
w^* = \arg \min_{w_i}  \sum_{i=1}^{K} \frac{M^i}{\lvert M \rvert} \mathcal{O}_i(D_i;w_i).    
\end{equation}
To this end, the model parameters are aggregated at the central server as:
\begin{equation}
w = \sum_{i=1}^{K} \alpha_i w_{i},
\end{equation}
where $\alpha_i$ is a hyperparameter controlling the importance of the local updates $w_i$ of $C_i$. The parameter transmission $w_i$ between the client and the central server causes the communication overhead. To reduce the communication overhead, we investigate the effectiveness of explanation-guided pruning, which requires an explanation method capable of attributing relevance not only to the input features but also to the model parameters. 
Backpropagation-based methods, such as LRP \cite{bach2015pixel} or Integrated Gradients \cite{sundararajan2017axiomatic}, are particularly suitable for this purpose. Given its proven success in the literature, we focus our attention on an LRP-based pruning method presented in \cite{yeom2021pruning, hatefi2024pruning}, and adapt it to an FL setup.

In LRP-based pruning, a DL model is viewed as a collection of $p$ interlinked components, denoted as  $\phi_i=\left\{\psi_1, \ldots, \psi_p\right\}$. 
These components can represent various structural elements, such as entire layers, groups of neurons, convolutional filters, or attention heads \cite{hatefi2024pruning}. The server has a global model $\phi^*$ and a reference set \( D_\text{ref} = \{(\boldsymbol{x}^z, \boldsymbol{y}^z)\}_{z=1}^{M} \), where \( M \) specifies the total number of reference samples.
To prune the global model's least important components, we compute the relevance $\bar{R}{\psi_c}$ of each component $\psi_c \in \phi^*$ over the reference set using LRP. For an input $I_z \in D_\text{ref}$ , LRP attributes relevance scores $R_k$ to each neuron $k$ by tracing its contribution to the output of the global model. The relevance $R_k^{(l)}$ in a layer $l$ is defined as follows:
\begin{equation}
R_k^{(l)}(I_z) = \sum_{j} \frac{v_{k,j}}{\sum_k v_{k,j}} R_j^{(l+1)}(I_z),
\end{equation}
where $v_{k,j}$ quantifies the contribution of neuron $k$ in layer $l$ to neuron $j$ in layer $l+1$, and $R_j^{(l+1)}$ represents the relevance of neuron $j$ in the subsequent layer. The relevance of neurons in the last layer $L$ is the logit output of the global model:
\begin{equation}
    R^{(L)}(I_z) = \phi(I_z).
\end{equation}
The relevance of a component $R_{\psi_c}\left(I_z\right)$ for an input $I_z \in D_\text{ref}$ is calculated by aggregating the relevance scores of the neurons that it contains. The overall relevance of a component in the global model $\bar{R}{\psi_c} $ is then estimated as the mean relevance over $D_{\text{ref}}$:
\begin{equation}
\label{eq:lrp_pru}
\bar{R}_{\psi_c} = \frac{1}{M} \sum_{z=1}^{M} R_{\psi_c}\left(I_z\right).
\end{equation}
The relevance scores for all components are collected into a set $\mathcal{R} = \{\bar{R}_{\psi_1}, \dots, \bar{R}_{\psi_p}\}$, which is sorted to identify the $q$ least relevant components, where $q$ is the pruning rate and determined as a percentage of total model parameters. Let $\bar{R}_{(\psi_1)} \le \bar{R}_{(\psi_2)} \le \cdots \le \bar{R}_{(\psi_p)}$ denote these values sorted in ascending order. The $q$-least relevant values are obtained as $ \mathcal{R}_q = \{\bar{R}_{(\psi_1)}, \bar{R}_{(\psi_2)}, \ldots, \bar{R}_{(\psi_q)}\}$.  To adapt the method for FL, we store the identified components in a pruning mask $\mathcal{M}$ as:
\begin{equation}
\label{eq:mask}
\mathcal{M}(\psi_c) = \begin{cases}
0 & \text{if } \bar{R}_{\psi_c} \in \mathcal{R}_q, \\[6pt]
1 & \text{otherwise.}
\end{cases}
\end{equation}
This mask is then applied to the global model as $w^* = w^* \odot \mathcal{M}$, eliminating irrelevant parameters. After the pruning, the central server distributes the sparse representation of the model to the clients. To minimize computational overhead, the pruning mask $\mathcal{M}$ is calculated only once. However, the pruning is repeated after each communication round using the existing mask for both the server and the clients. This ensures consistency across models and further reduces the communication overhead. The iterative pruning process begins after a warmup phase of $\upsilon$ communication rounds, ensuring the global model is robust enough to distinguish relevant from irrelevant components before removing the least relevant ones. 
The model distribution is illustrated in Fig. \ref{figure:pruning}.

\section{Experimental Results}

In the experiments, we assessed our strategy in the context of multi-label image classification in RS. We conducted the experiments using the BigEarthNet-S2 v2.0 benchmark archive \cite{clasen2024reben}. A subset of BigEarthNet-S2, comprising images collected during the summer season from Austria, Belgium, Finland, Ireland, Lithuania, Serbia, Portugal, and Switzerland, was selected. Each image is annotated with multi-labels derived from the CORINE Land Cover Map database, following the 19-class nomenclature defined in \cite{sumbul2021bigearthnet}. 
 We used the train-test split recommended in \cite{clasen2024reben}. Furthermore, we allocated the training data such that each client exclusively held data from a single country.

We used FedAvg \cite{li2019convergence} as the FL algorithm, although any FL algorithm could be used. The FedAvg algorithm updates the global model by aggregating the weighted averages of locally trained models from the distributed clients. The results of FedAvg with our strategy are denoted as \textit{proposed} and compared with: 1) those obtained by using FedAvg without pruning (denoted as \textit{standard}); and 2) those obtained by using FedAvg with random pruning (denoted as \textit{random}). We varied the pruning rate from 10\% to 40\% with an increment of 10\%. To maintain a balance between communication overhead and generalization capability, we limit the maximum pruning rate to 40\%, which provides meaningful compression without severely affecting model generalization. We selected a ResNet-50 architecture as the base model for our experiments due to its good performance and widespread usage in image classification tasks. The model was trained for 20 communication rounds with 3 local training epochs per round. After 20 communication rounds, we observed that the generalization capability of the model did not improve significantly. The number of clients was set to 8, all participating in every communication round. We chose the Adam optimizer with a learning rate of 0.001 and a mini-batch size of 512. The warmup phase ended after 9 communication rounds ($\upsilon=9$). The LRP hyperparameters for ResNet-50 were adapted directly from \cite{hatefi2024pruning}. We assessed performance based on MLC accuracy, measured by the mean Average Precision (mAP) metric.

\begin{table}[t]
\renewcommand{\arraystretch}{1.5}
\setlength\tabcolsep{6pt}
\caption{mAP scores (\%) obtained by the standard, random, and proposed strategies considering different pruning rates after 20 communication rounds.}
\label{tab:T_1}
\centering
\begin{tabular}{cccccc}
\hline
\multirow{2}{*}{\textbf{Strategies}} & \multicolumn{5}{c}{\textbf{Pruning Rate (\%)}}                                                                                                   \\ \cline{2-6} 
                           &{\textbf{0}} & {\textbf{10}} & {\textbf{20}} & {\textbf{30}} &
                           {\textbf{40}} 
                             \\
 \hline
 Standard& {65.40}& -& -& -&-\\ \hline

Random&-& {68.49}& {66.64}& {66.40}& {65.20}\\ \hline
Proposed& -& {67.47} & {68.78}& {67.90}& {68.53}\\ \hline     
\end{tabular}
\end{table}
\begin{figure}[h]
    \centering
    \includegraphics[width=1\linewidth]{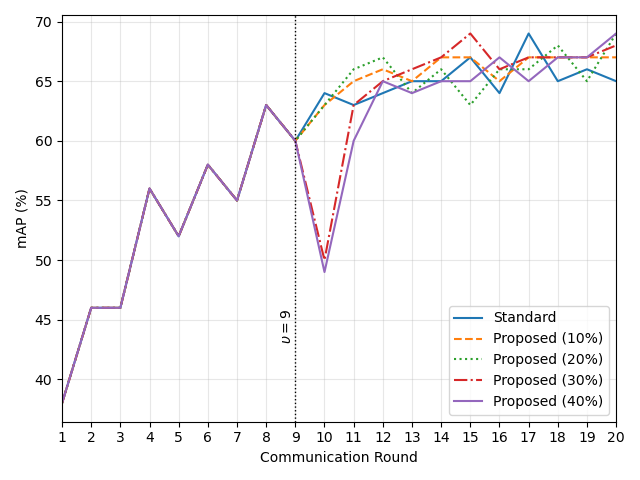}
    \caption{mAP scores (\%) versus communication rounds obtained by the standard and the proposed strategies, considering different pruning rates.}
    \label{fig:training+graph}
\end{figure}
Table \ref{tab:T_1} shows the corresponding results for the standard, random, and proposed strategies at different pruning rates. From the table, one can observe that the proposed strategy achieves higher mAP scores than the standard strategy across all pruning rates. As an example, our strategy with a 20\% pruning rate achieves a 3.38\% higher mAP score compared to the standard strategy. This indicates that our strategy is capable of pruning less important global model parameters, enabling the global model to focus on relevant features, and thereby reducing overfitting. By analyzing the table, we can also observe that our strategy outperforms random pruning at different pruning rates. Compared to random pruning, the proposed strategy shows clear improvements at pruning rates of 20\%, 30\%, and 40\%, with gains of 2.14\%, 1.50\%, and 3.33\%, respectively. This shows that our strategy removes parameters that have less impact on the performance of the global model. However, random pruning outperforms our strategy at a pruning rate of 10\%. This could be due to the global model retaining highly redundant or irrelevant information, leading to suboptimal performance at low pruning rates.

Fig. \ref{fig:training+graph} shows the mAP scores of our strategy with different pruning rates and the standard strategy across different communication rounds. Until the ninth round (when pruning is applied), the performance is identical for both strategies. Pruning a large number of parameters in the ninth communication round leads to a significant drop in the mAP score with our strategy. This decline is particularly visible at pruning rates of 30\% and 40\%, where a higher proportion of parameters are removed. As the pruned model parameters are less informative, the mAP score is increased to the performance of the standard strategy within a few rounds after the pruning round. As an example, in the thirteenth communication round, the standard strategy and our strategy with various pruning rates achieve similar mAP scores. As the number of communication rounds increases, our strategy outperforms the standard strategy at all pruning rates. This shows that even when we reduce the model parameters shared with the global server by 40\%, we achieve a higher mAP score compared to using all the model parameters. Additionally, we would like to highlight that the computational cost of training a global model is almost identical to that of the standard strategy. The average completion time for a training round increases by only 2\% when using our strategy compared to the standard strategy. This indicates that our strategy not only improves the generalization capability of the global model while reducing the number of shared model parameters and avoiding additional computational costs.

\section{Conclusion}

In this paper, we have introduced an explanation-guided pruning strategy to reduce communication overhead in FL in the context of RS image classification. By leveraging LRP-based pruning, our strategy identifies and removes less informative model parameters, and thus reduces the communication overhead between clients and the central server. Since most of the computations take place on the central server, it avoids significant computational demands on the clients. We would like to emphasize that our strategy is independent of the number of clients, FL architectures, learning tasks, and FL aggregation algorithms. Experimental results on the BigEarthNet-S2 benchmark demonstrate that the proposed strategy reduces communication overhead, while improving the global model's generalization capability compared to standard FL training and random pruning strategies. We would like to note that the proposed strategy assumes that the clients are associated with equal importance during model aggregation. However, in real applications, different clients may have different levels of importance. In future work, we plan to investigate the effectiveness of the explanation methods for the estimation of the importance scores for the clients, aiming to enable more effective aggregation of their contributions to the global model.    

\bibliographystyle{IEEEtran}
{\small

\bibliography{main}}
\end{document}